\documentclass[sigconf, nonacm, review=False, timestamp=False]{acmart}
\usepackage{booktabs}
\usepackage{multirow}
\usepackage{graphicx}
\usepackage{float}
\AtBeginDocument{%
  \providecommand\BibTeX{{%
    \normalfont B\kern-0.5em{\scshape i\kern-0.25em b}\kern-0.8em\TeX}}}

\setcopyright{acmcopyright}
\copyrightyear{2022}
\acmYear{2022}
\acmDOI{XXXXXXX.XXXXXXX}

\acmConference[AIES]{5th AAAI/ACM Conference on AI, Ethics, and Society}{August 01--03, 2022}{Oxford, UK}
\acmPrice{15.00}
\acmISBN{978-1-4503-XXXX-X/18/06}

\begin{document}
\graphicspath{{.}{./figures/}}
\title[Write It Like You See It]{Write It Like You See It: Detectable Differences in Clinical Notes By Race Lead To Differential Model Recommendations}

\author{Hammaad Adam}
\email{hadam@mit.edu}
\orcid{0000-0001-6910-7074}
\affiliation{%
  \institution{Massachusetts Institute of Technology}
  \city{Cambridge}
  \state{MA}
  \country{USA}
  \postcode{02139}
}
\author{Ming Ying Yang}
\email{ming1022@mit.edu}
\affiliation{%
  \institution{Massachusetts Institute of Technology}
  \city{Cambridge}
  \state{MA}
  \country{USA}
  \postcode{02139}
}

\author{Kenrick Cato}
\email{kdc2110@cumc.columbia.edu}
\affiliation{%
  \institution{Columbia University School of Nursing}
  \city{New York}
  \state{NY}
  \country{USA}
  \postcode{10032}
}

\author{Ioana Baldini}
\email{ioana@us.ibm.com}
\affiliation{%
  \institution{IBM Research}
  \city{Yorktown}
  \state{NY}
  \country{USA}
  \postcode{10598}
}

\author{Charles Senteio}
\email{charles.senteio@rutgers.edu}
\affiliation{%
  \institution{Rutgers University School of Communication and Information}
  \city{New Brunswick}
  \state{NJ}
  \country{USA}
  \postcode{08901}
}

\author{Leo Anthony Celi}
\email{lceli@mit.edu}
\affiliation{%
  \institution{Massachusetts Institute of Technology}
  \city{Cambridge}
  \state{MA}
  \country{USA}
  \postcode{02139}
}

\author{Jiaming Zeng}
\email{jiaming@ibm.com}
\affiliation{%
  \institution{IBM Research}
  \city{Cambridge}
  \state{MA}
  \country{USA}
  \postcode{02139}
}

\author{Moninder Singh}
\email{moninder@us.ibm.com}
\affiliation{%
  \institution{IBM Research}
  \city{Yorktown}
  \state{NY}
  \country{USA}
  \postcode{10598}
}
\author{Marzyeh Ghassemi}
\email{mghassem@mit.edu}
\affiliation{%
  \institution{Massachusetts Institute of Technology}
  \city{Cambridge}
  \state{MA}
  \country{USA}
  \postcode{02139}
}








\renewcommand{\shortauthors}{ }

\begin{abstract}
Clinical notes are becoming an increasingly important data source for machine learning (ML) applications in healthcare. Prior research has shown that deploying ML models can perpetuate existing biases against racial minorities, as bias can be implicitly embedded in data. In this study, we investigate the level of implicit race information available to ML models and human experts and the implications of model-detectable differences in clinical notes. Our work makes three key contributions. First, we find that models can identify patient self-reported race from clinical notes even when the notes are stripped of explicit indicators of race. Second, we determine that human experts are not able to accurately predict patient race from the same redacted clinical notes. Finally, we demonstrate the potential harm of this implicit information in a simulation study, and show that models trained on these race-redacted clinical notes can still perpetuate existing biases in clinical treatment decisions.
\end{abstract}

\maketitle

\section{Introduction}
There are a number of well-established inequities in hospital-based healthcare delivery that affect patients from racial minority groups. Minority patients have worse hospital outcomes across various medical conditions \cite{27-Hebert2017-hf}, including congestive heart failure \cite{4-Alexander1995-tj}, myocardial infarction \cite{53-Skinner2005-rg}, and perinatal care \cite{29-Howell2013-ft}, as well as for various surgical procedures \cite{52-Silber2009-jm}. Minority patients tend to receive care from different physicians than non minority patients; the physicians they see have less clinical training \cite{53-Skinner2005-rg}, are less likely to be board certified, and are more likely to report that they are unable to provide high-quality care to all their patients \cite{8-Bach2004-xb}. These structural factors along with physician implicit biases create inequitable treatment decisions \cite{25-Green2007-rn}; for example, physicians are less likely to provide Black patients with analgesia for acute pain in the emergency room \cite{33-Lee2019-bf} or thrombolysis for acute coronary syndromes \cite{25-Green2007-rn}.

These disparities are coming into increasing focus as the use of machine learning (ML) and artificial intelligence (AI) proliferates in healthcare. Deploying ML models in clinical settings has been proposed to improve diagnostic accuracy, treatment decisions, and operational efficiency \cite{55-Topol2019-fc}. However, the use of models also risks replicating and exacerbating implicit biases present in the data used to train them. Prior research found that health systems were far less likely to refer Black patients to high-risk care management programs than similar White patients, because they relied on algorithms that used healthcare costs as a proxy for health \cite{42-Obermeyer2019-oi}. Similar inequities have been described when ML models have been deployed in other high-stakes domains, including criminal justice \cite{6-Angwin2016-gf} and financial lending \cite{39-Martinez2021-gr}.
 
Notably, ML models can perpetuate existing biases even if they do not have explicit access to race. Models that were less likely to recommend Black patients to high-risk care management programs \cite{42-Obermeyer2019-oi}, more likely to identify Black defendants as high risk \cite{6-Angwin2016-gf}, and less likely to approve Black mortgage applicants \cite{39-Martinez2021-gr} all did not explicitly use race as a variable in making their predictions. However, the models were able to infer race from other correlated attributes, and could thus propagate existing human biases in these decisions. To construct fair models, it is thus vital for machine learning practitioners to understand and account for the implicit racial information present in the data they train their models on. Such information is not always obvious, and can exist in data sources that seem race-redacted to humans \cite{9-Banerjee2021-zw}. However, little work has focused on investigating the level of race information contained in clinical notes \cite{14-Boag2018-gx}, despite these being an increasingly common source of data for ML models \cite{14-Boag2018-gx,20-Ghassemi2014-bt,34-Lehman2012-wp, 63-Zeng2022-fz}.
 
In this study, we investigate the presence of implicit racial information in clinical nursing notes that have had direct race information redacted. Using data from two large hospitals, we determine that models are able to accurately predict a patient’s self-reported race from notes written during their hospital stay, even after explicit indicators of race are removed. We conduct a detailed audit of the drivers of predictive accuracy to identify potential disparities in clinical care, and find that while some variations are clinically justifiable (e.g. comorbidities that are more common in Black patients), others may reflect potential avenues of missed care (e.g. references to bruising and rashes being extremely predictive of White race, even though there is no clinical reason for these symptoms to be less common in Black patients). Notably, human experts do not share the ability to identify race: a group of 42 surveyed physicians were unable to accurately determine patient race using the same redacted clinical notes. While the ability of models to predict race is not intrinsically harmful, it implies that ML models trained on clinical notes have access to patient race, even if this information is not explicitly provided and undetectable by clinicians. We illustrate the potential harm of this information in a simulation study, demonstrating that models trained on these race-redacted clinical notes perpetuate existing biases in clinical treatment decisions.

\section{Related Work}

We discuss three categories of prior research that are directly related to and impacted by our findings: studies that attempt to predict race from clinical data, audits of racial biases in clinical notes, and clinical prediction tasks.

\textit{Race Prediction From Clinical Data.} Relatively little research has focused on characterizing the implicit racial information present in clinical data by attempting to predict patient race. One such study focused on medical images, and revealed that deep learning algorithms were able to accurately predict self-reported patient race exclusively from chest X-rays, though expert radiologists were completely unable to \cite{9-Banerjee2021-zw}. To our knowledge, only one other study has attempted such a task with clinical notes. This paper used the publicly available MIMIC dataset \cite{30-Johnson2016-fo}, and found that while age and gender were easy to predict from clinical notes, race posed a harder challenge \cite{14-Boag2018-gx}. Their final model was only able to distinguish between White and non-White patients with an area under the receiver operator curve (AUC) of 0.62. However, their analysis grouped all non-white minorities into a single combined category, which severely limited predictive accuracy. In contrast, we focus on differentiating between Black and White patients; this approach avoids grouping heterogeneous populations \cite{60-Williams2010-fk} and allows us to specifically focus on disparities faced by Black patients. We have not found any work that attempts to assess the ability of human experts to predict race from clinical notes. 

\textit{Racial Biases in Clinical Notes.} Recent work has established that clinical notes may not be written in the same way for all types of patients, and may reflect racial disparities in clinical care. For instance, notes written for Black patients are much more likely to contain indicators of physician mistrust \cite{10-Beach2021-od}, negative descriptors \cite{54-Sun2022-bh}, and other stigmatizing language \cite{45-Park2021-tw} than those written for White patients. Clinical notes have revealed that Black patients have lower levels of trust in their physicians during end of life care \cite{13-Boag2018-tp}. These studies all audit a specific bias in clinical notes by identifying language that varies significantly by patient race. Our work adopts a broader approach, and characterizes all differences in content and language that are predictive of race.

\textit{Clinical Prediction Tasks.} Clinical notes are becoming an increasingly common data source for machine learning applications in healthcare. Several studies have used ML models to predict patient outcomes such as in-hospital mortality \cite{20-Ghassemi2014-bt, 26-Grnarova2016-gf, 34-Lehman2012-wp, 62-Ye2020-jw}, 30-day mortality \cite{36-Luo2016-aj}, and readmission \cite{24-Golmaei2021-bq, 34-Lehman2012-wp, 51-Rumshisky2016-iu}. Language models like MedBERT \cite{48-Rasmy2021-xf} and Clinical BERT \cite{5-Alsentzer2019-qe} have also demonstrated excellent performance on tasks like disease prediction, de-identification, and named entity recognition. However, recent work has demonstrated that such models can exhibit performance gaps for racial and gender subgroups \cite{64-Zhang2020-ze}. Our work emphasizes the dangers of naively using such models for clinical prediction. We demonstrate that even if ML models are trained on seemingly race-redacted data, they may still propagate existing biases in clinical decisions.

\section{Data}

Our dataset consists of clinical notes from two sites: Beth Israel Deaconess Medical Center in Boston and Columbia University Medical Center In New York. De-identified clinical notes from Beth Israel are publicly available through the Medical Information Mart for Intensive Care III (MIMIC-III) version 1.4 database \cite{30-Johnson2016-fo}. The notes from Columbia are private data available from electronic health records (EHR), and contain protected health information (PHI). The use of this data was approved by Columbia’s Institutional Review Board (IRB). 

In our analyses, we focused on progress updates and other clinical notes written by nurses. In addition to these nursing notes, we extracted a patient’s self-reported race and other demographic information from their EHR. We only included patients who self-report their race as White/Caucasian and Black/African-American due to smaller numbers of other self-reported race categories. To ensure similarity between the two datasets, our analysis only included adult patients (i.e. over 18) admitted to non-pediatric units, as well as infants admitted to the neonatal ICU (NICU). For patients admitted multiple times, we only considered their first stay.

Our final dataset contained 668,768 notes written for 28,032 patients from MIMIC and 3,554,802 notes for 29,807 patients from Columbia. Table \ref{tab:s1-demographic} in the Appendix presents a summary of the resulting cohort’s demographics. We note that many existing works in clinical natural language processing (NLP) rely exclusively on the publicly available MIMIC dataset. Our ability to analyze data from both MIMIC and the Columbia datasets is important, especially as the patient populations differ significantly across the two sites. Most notably, Columbia sees a much higher proportion of Black patients (${\sim}20\%$ vs. ${\sim}10\%$).

Before conducting our analyses, we redacted any explicit mentions of patient race from the nursing notes in both datasets. We compiled a list of terms that were used as identifiers of race, and removed these from the two corpora of notes using regular expression operations. These terms were identified by training a logistic regression classifier to predict race using a unigram bag-of-words (BoW) representation of the notes in both datasets. We manually inspected the most predictive terms for each race (according to the model’s coefficients), and identified all terms that could be used as an explicit indicator of patient race. The final list contained the terms African-American, African, Black, and Creole as identifiers of Black race, and the words Caucasian and White as identifiers of White race. These terms were removed regardless of capitalization (e.g. black vs Black), case (e.g. AFRICAN vs African), and hyphenation (e.g. African-American vs. African American). As the Columbia dataset contains PHI, we also removed mentions of area codes, neighborhoods, and hospitals that served as proxies for race. A full list of these terms is provided in the appendix.

\section{Model-Based Race Detection}

The goal of our primary analysis is to demonstrate that ML models are able to identify a patient’s self-reported race from nursing notes that describe their condition and progress. We find that even after nursing notes are stripped of explicit racial identifiers, models are able to accurately predict patient race. This accurate prediction was found across different sites, units, and patient types. Investigating the drivers of predictive performance, we determined that clinical notes written for White and Black patients vary greatly in content. While some of the identified differences are clinically justifiable, others might suggest disparities in clinical care and warrant further investigation. 

\subsection{Methods}
\begin{figure*}[!t]
    \centering
    \includegraphics[width=.7\textwidth]{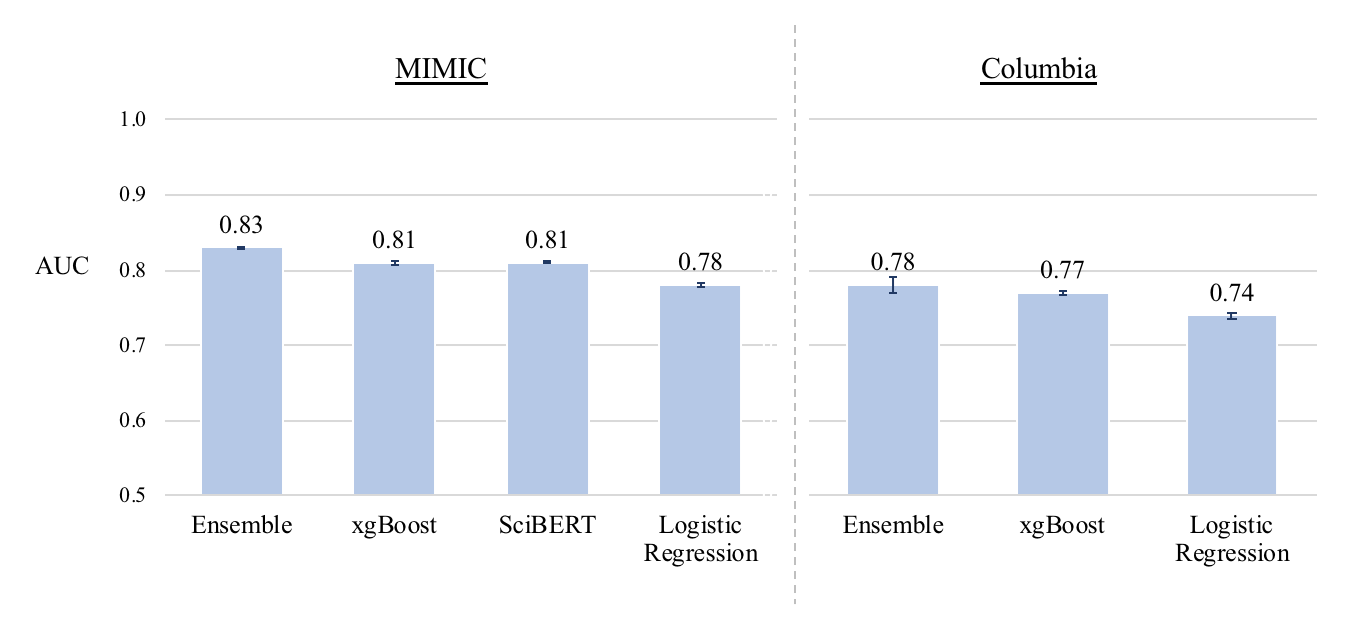}
    \caption{Model classification performance for patient self-reported race from nursing notes. The chart displays the mean AUC (across 10 random train-test splits) and error bars that signify 95\% confidence intervals. The ensemble of xgBoost and logistic regression classifiers demonstrate the highest accuracy in both datasets.}
    \label{fig:mimic + colombia AUC}
\end{figure*}

We trained four machine learning models to predict a patient’s self-reported race from nursing notes written during their stay. Crucially, these models were trained on notes that were stripped of any explicit indicators of patient race. We used a unigram\footnote{We also tried bigram representations, but found that these did not boost performance. See Appendix \ref{sec:bigram}} BoW representation of the nursing notes to train an L1-penalized logistic regression \cite{46-Pedregosa2011-yz} and an xgBoost \cite{17-Chen2016-jk} classifier, as well as a stacking ensemble of the two methods. These three models performed their prediction at the visit level: all notes written for the same patient were aggregated into one large note that was then used for prediction. This approach allowed us to easily pool all the available information on each patient, and yielded higher predictive accuracy. 

In addition to the bag-of-words models, we also fine-tuned SciBERT \cite{12-Beltagy2019-gu}–a language model trained on scientific abstracts–to classify patient self-reported race (note that we did not use clinical note-specific models like Clinical BERT \cite{5-Alsentzer2019-qe}, as some of these have already been trained on the MIMIC data, creating possible information leakage). For SciBERT, we first fine-tuned the model to predict patient self-reported race from individual nursing notes (as opposed to one combined note), then aggregated the predictions by patient. In the aggregation step, we considered the model to have predicted a patient’s race as Black if it did so for any of their individual notes. This approach was necessitated by the fact that the combined notes were usually much longer than SciBERT’s 512 token limit. Note that we were unable to test the SciBERT classifier on the Columbia dataset due to privacy and computational constraints (the data contains PHI, and must stay on a private server that does not have access to a GPU).

\begin{table}[!b]
\centering
\caption{Detailed classification accuracy achieved by the ensemble method in predicting race from nursing notes. We evaluated the classifier on ten random train-test splits, and report the mean and standard deviation of test set AUCs across splits. We report accuracy on the whole test set population, as well as patient subgroups based on specific comorbidities, VW comorbidity score \cite{58-Van_Walraven2009-wg} decile, and unit type.}
\label{tab:auc by group}
\resizebox{.7\linewidth}{!}{%
\begin{tabular}{@{}llcc@{}}
\toprule
\multicolumn{1}{c}{{\color[HTML]{212121} }} &
  \multicolumn{1}{c}{{\color[HTML]{212121} }} &
  \multicolumn{2}{c}{AUC} \\
\multicolumn{1}{c}{\multirow{-2}{*}{{\color[HTML]{212121} Subgroup}}} &
  \multicolumn{1}{c}{{\color[HTML]{212121} }} &
  {\color[HTML]{212121} MIMIC} &
  {\color[HTML]{212121} Columbia} \\ \midrule
{\color[HTML]{212121} Overall} &
  {\color[HTML]{212121} } &
  0.83 (0.00) &
  0.78 (0.02) \\
\multicolumn{1}{c}{{\color[HTML]{212121} }} &
  \multicolumn{1}{c}{{\color[HTML]{212121} }} &
   &
   \\
Comorbidity               &             &             &             \\
\multicolumn{2}{l}{Diabetes}            & 0.80 (0.00) & 0.76 (0.01) \\
\multicolumn{2}{l}{Hypertension}        & 0.82 (0.01) & 0.80 (0.02) \\
\multicolumn{2}{l}{COPD}                & 0.82 (0.02) & 0.79 (0.00) \\
\multicolumn{2}{l}{Obesity}             & 0.77 (0.00) & 0.76 (0.02) \\
                          &             &             &             \\
Comorbidity   Score       &             &             \\
\multicolumn{2}{l}{Top decile}          & 0.83 (0.00) & 0.73 (0.01) \\
\multicolumn{2}{l}{Bottom decile}       & 0.81 (0.01) & 0.80 (0.02) \\
                          &             &             &             \\
Unit                      &             &             &             \\
\multicolumn{2}{l}{MICU}                & 0.81 (0.00) & 0.75 (0.01) \\
\multicolumn{2}{l}{CCU}                 & 0.83 (0.03) & 0.82 (0.01) \\
\multicolumn{2}{l}{NICU}                & 0.81 (0.01) & 0.71 (0.00) \\
\multicolumn{2}{l}{SICU}                & 0.80 (0.01) & 0.76 (0.01) \\
\multicolumn{2}{l}{TSICU}               & 0.80 (0.02) & -           \\
\multicolumn{2}{l}{NUICU}               & -           & 0.78 (0.00) \\
\multicolumn{2}{l}{CSRU}                & -           & 0.75 (0.00) \\ \bottomrule
\end{tabular}%
}
\end{table}

\begin{figure*}[!t]
    \centering
    \includegraphics[width=0.7\textwidth]{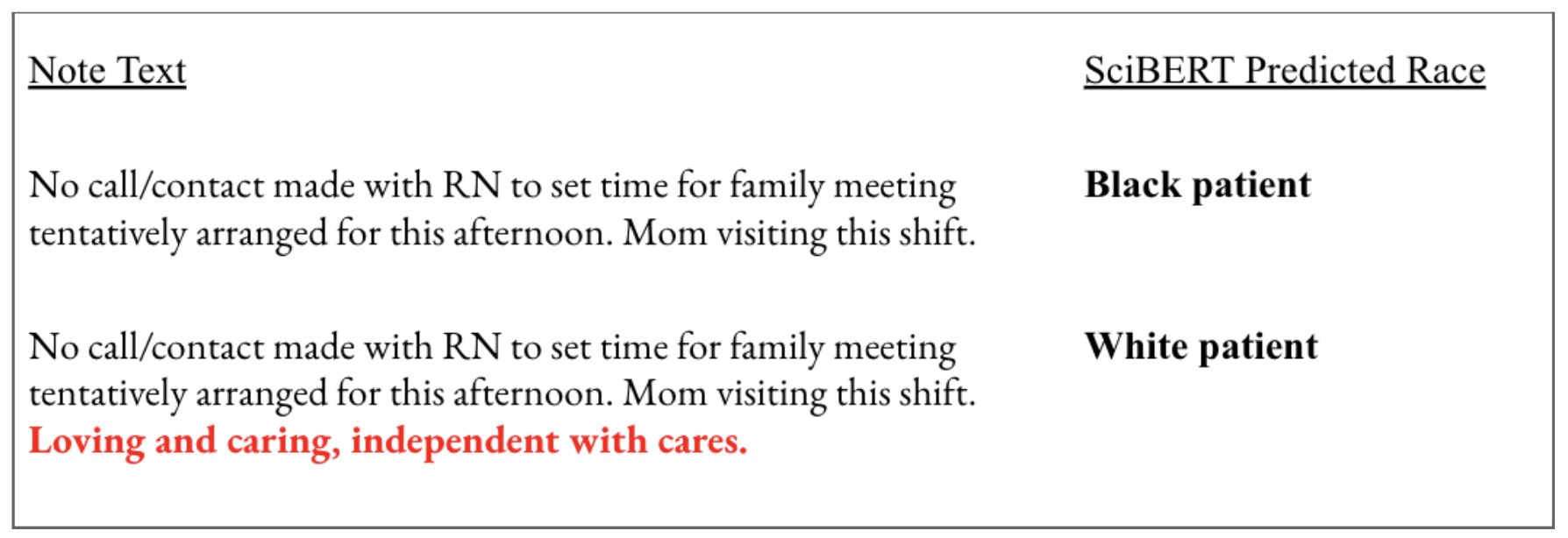}
    \caption{An adversarial example demonstrating the association between White race and positive descriptors in the SciBERT model’s predictions. We take an excerpt of a note written for a Black patient (in black text), and an excerpt of a note written for a White patient (in red text). Both excerpts were taken from notes from the test set (i.e. not the data the model was trained on) that the model predicted correctly. Adding a positive descriptor of the patient’s family led the model to change its prediction of the patient’s race from Black to White.}
    \label{fig:example note}
\end{figure*}

We trained and evaluated these models on ten random 7:3 train-test splits, evaluating performance by the area under the receiver operator curve (AUC) on the test set of patients after models are trained to convergence on training data only. We then inspected models in two ways to uncover the differences in the content of notes written for Black and White patients. First, we examined which coefficients of the logistic regression were most predictive of patient race, specifically working to identify how the words predictive of Black race differ from those predictive of White race. Second, we ran a structural topic model (STM) \cite{48-Rasmy2021-xf} to identify more nuanced differences between notes written for Black and White patients in an unsupervised manner. 

\subsection{Results}
\subsubsection{Self-Reported Race is Predictable from Nursing Notes.}
We found that a patient’s self-reported race is predictable from nursing notes written during their hospital stay, even after explicit indicators of race are removed. All models were able to distinguish between Black and White patients (Figure \ref{fig:mimic + colombia AUC}), with the best model achieving 0.83 AUC on the MIMIC dataset and 0.78 AUC on the Columbia dataset. Crucially, predictive performance is not driven by a specific characteristic or patient type: the results are consistent across various subgroups of the held out set, including patients with diabetes, hypertension, chronic pulmonary disease, and obesity, as well as patients admitted to different units (Table \ref{tab:auc by group}). While there is some variation in performance, the models all perform above 0.7 AUC on all categories in both datasets. The fact that race is predictable in two large hospitals in different cities with very different patient populations is notable, and speaks to the generalizability of our primary result.

\subsubsection{Notes Written for Black and White Patients Differ Significantly} 
The finding that an algorithm can distinguish between White and Black patients is not troubling on its own. For example, Black populations have higher rates of comorbidities like diabetes, asthma, and obesity \cite{19-Forno2009-cv, 31-Kabarriti2020-md}; their notes are likely to mention these conditions more often, creating a pattern that an algorithm will be able to pick up on. However, a similar pattern could also be created by a different standard of care for White and Black patients \cite{54-Sun2022-bh}, which is more concerning. For instance, if stigmatizing language and negative descriptors are used more frequently for Black patients \cite{45-Park2021-tw, 54-Sun2022-bh}, models would also be able to rely on such associations to identify patient race. 

The SciBERT model trained to predict self-reported race exhibits such a trend, as it associated positive descriptors of patient family with White race. In several instances, adding a positive description of the patient’s family to the note led SciBERT to change its prediction of the patient’s race from Black to White (Figure \ref{fig:example note}). This suggests that the model may have learnt an association between “loving and caring” and White race. 

\begin{figure*}[h]
    \centering
    \includegraphics[width=0.8\textwidth]{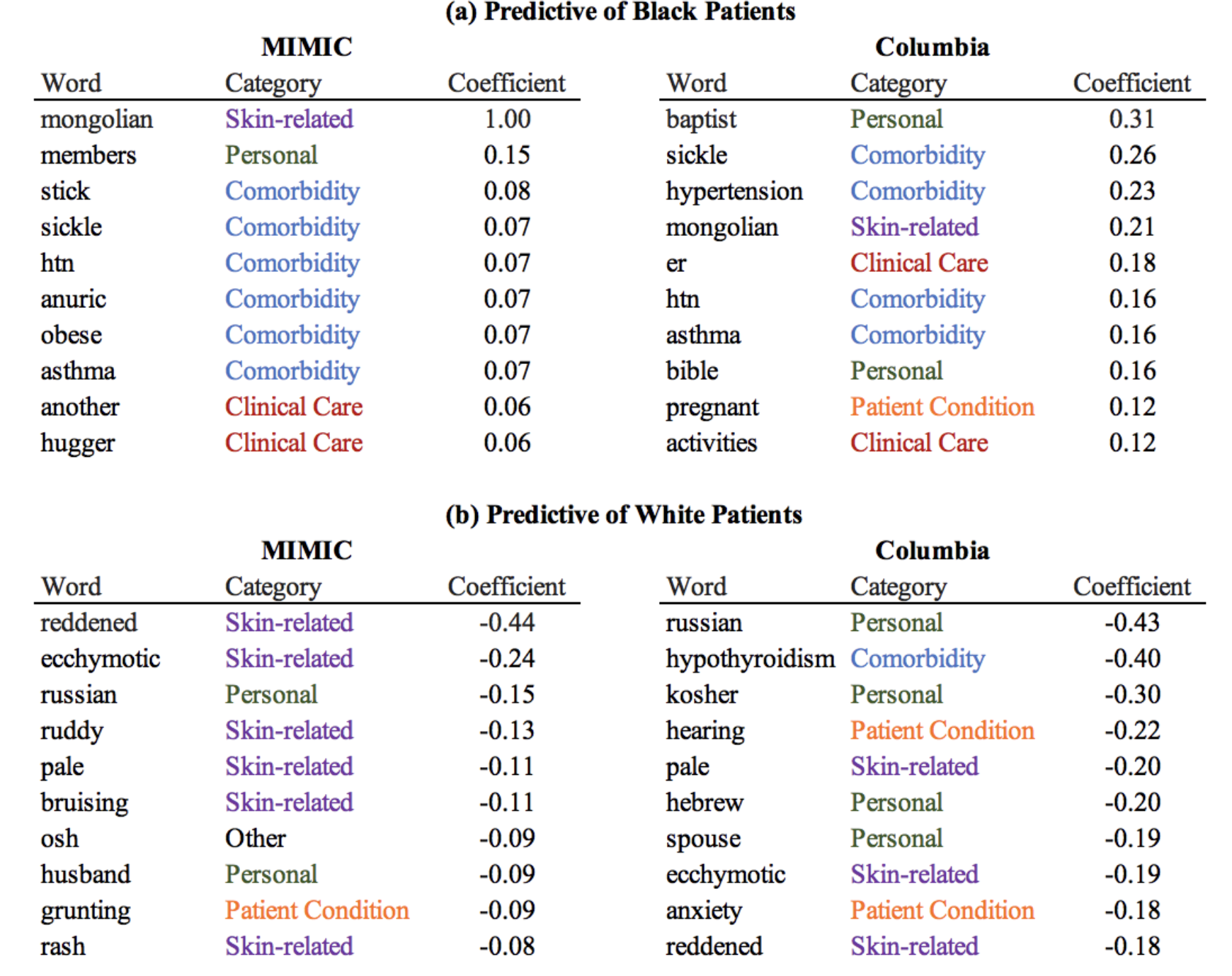}
    \caption{Words that are most predictive of race in nursing notes, sorted by the word’s logistic regression coefficient. We categorized the predictive words into five clinically motivated categories: skin-related, personal, comorbidity, clinical care, and patient condition.}
    \label{fig:top words}
\end{figure*}

We investigated the 25 words most predictive of each racial group (on average across train-test splits), classifying them into five clinically motivated categories: skin-related, personal, comorbidity, clinical care, and patient condition (Figure \ref{fig:top words}). We find that Black patients are often identified by comorbid conditions like sickle cell anemia, asthma, and diabetes, which are more common in Black patients \cite{23-Glassberg2013-ls, 31-Kabarriti2020-md, 56-Tschudy2016-ub}. However, references to skin like bruising, redness, or paleness are strong predictors of White self-reported race, but these words don’t necessarily reflect conditions that should be more common for White patients. Paleness, redness, and bruising are all clinical symptoms that should be noted for both White and Black patients. The fact that they are strongly associated with White skin is troubling in the context of previous work that suggests that healthcare providers are less equipped to diagnose skin conditions in patients with darker skin. A number of reviews have found that only a small fraction of examples provided in dermatology textbooks are on non-white skin, which can lead to serious underdiagnosis \cite{2-Adelekun2021-hg, 35-Lester2020-sm}.

\begin{table}[!b]
\centering
\caption{References to family for married, female patients in MIMIC. The table displays the percentage of patients by race whose notes contain at least one mention of the given word. Personal descriptors like “husband” and “father” are more common in notes written for White patients, while the group descriptor “family members” is more common in notes written for Black patients.}
\label{tab:racial diff}
\resizebox{.75\linewidth}{!}{%
\begin{tabular}{@{}lrr@{}}
\toprule
Word                        & \% Black Patients & \% White Patients  \\ \midrule
husband                     & 50\%              & 64\%\\
family members              & 22\%              & 14\% \\
father                      & 2\%               & 5\%   \\ \bottomrule
\end{tabular}%
}
\end{table}

\begin{figure*}[htp]
    \centering
    \includegraphics[width=0.7\textwidth]{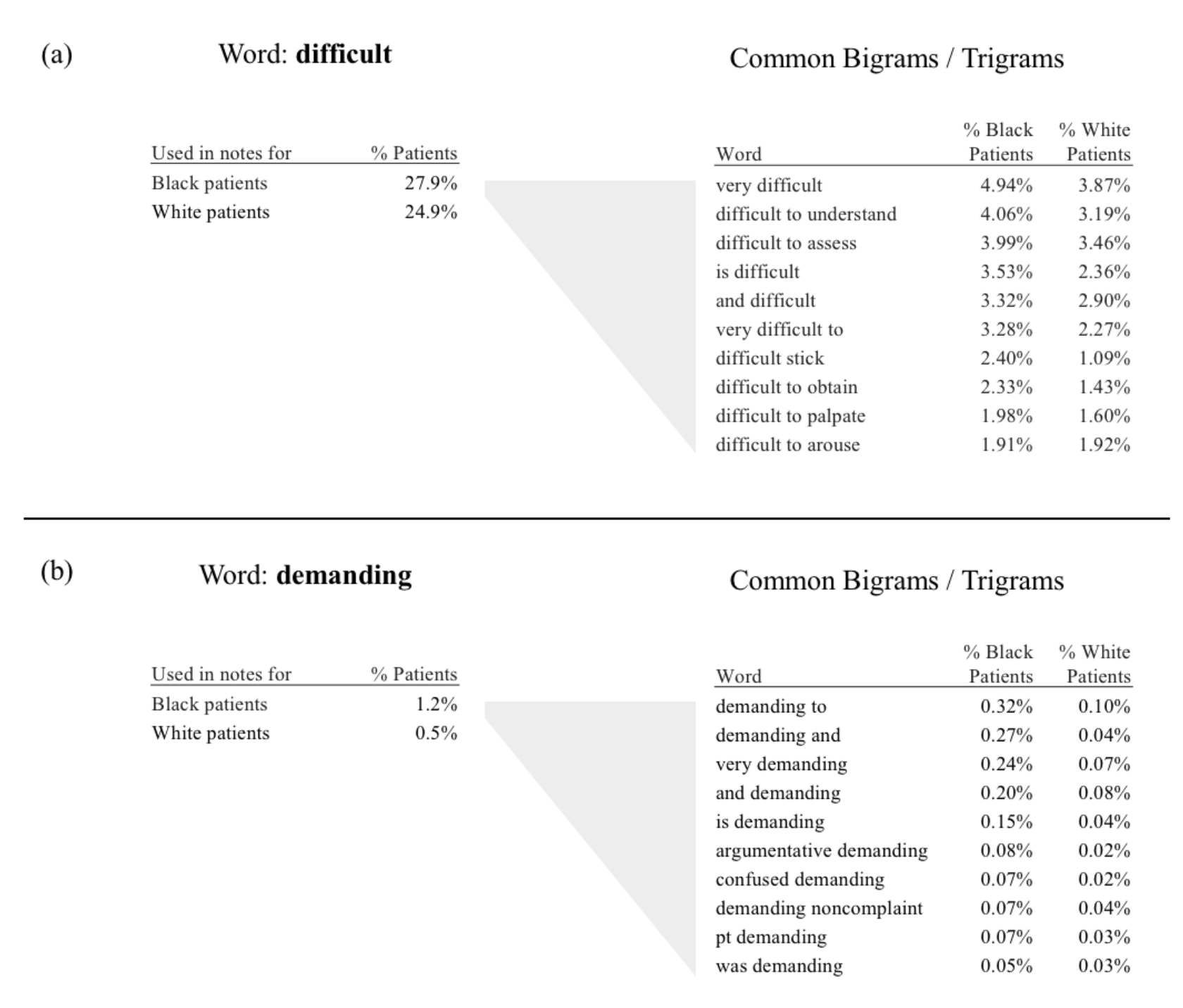}
    \caption{Common bigrams and trigrams for the word difficult in the MIMIC dataset and demanding in the Columbia dataset, both of which are predictive of Black race. We measure the frequency of each term as the percentage of patients of a given race whose notes contained the term at least once.}
    \label{fig:bigram trigram}
\end{figure*}

We also find differences in words that may be subjective rather than clinical. For instance, the phrase “family members” is associated with Black patients in MIMIC, while “husband” and “father” are associated with White patients. Some of this trend can be explained by population differences: more White women in the sample are married. However, even if we only consider married female patients, husband is still referred to more often for White patients than Black (Table \ref{tab:racial diff}). Another example is that the word “difficult” is predictive of Black race in the MIMIC dataset, while “demanding” is predictive of Black race in the Columbia dataset. As Figure \ref{fig:bigram trigram} demonstrates, this word can be used in many contexts: saying a patient is difficult is very different from saying that they are a “difficult stick” (i.e. a patient whose veins are hard to insert a needle into). While the latter is a more objective claim, the first is subjective, and may hint at differential treatment. Statements implying that the patient was “very difficult” or “very demanding” were more frequent for Black patients, which is a concerning trend. (Figure \ref{fig:bigram trigram})

\begin{figure*}[!t]
    \centering
    \includegraphics[width=.8\textwidth]{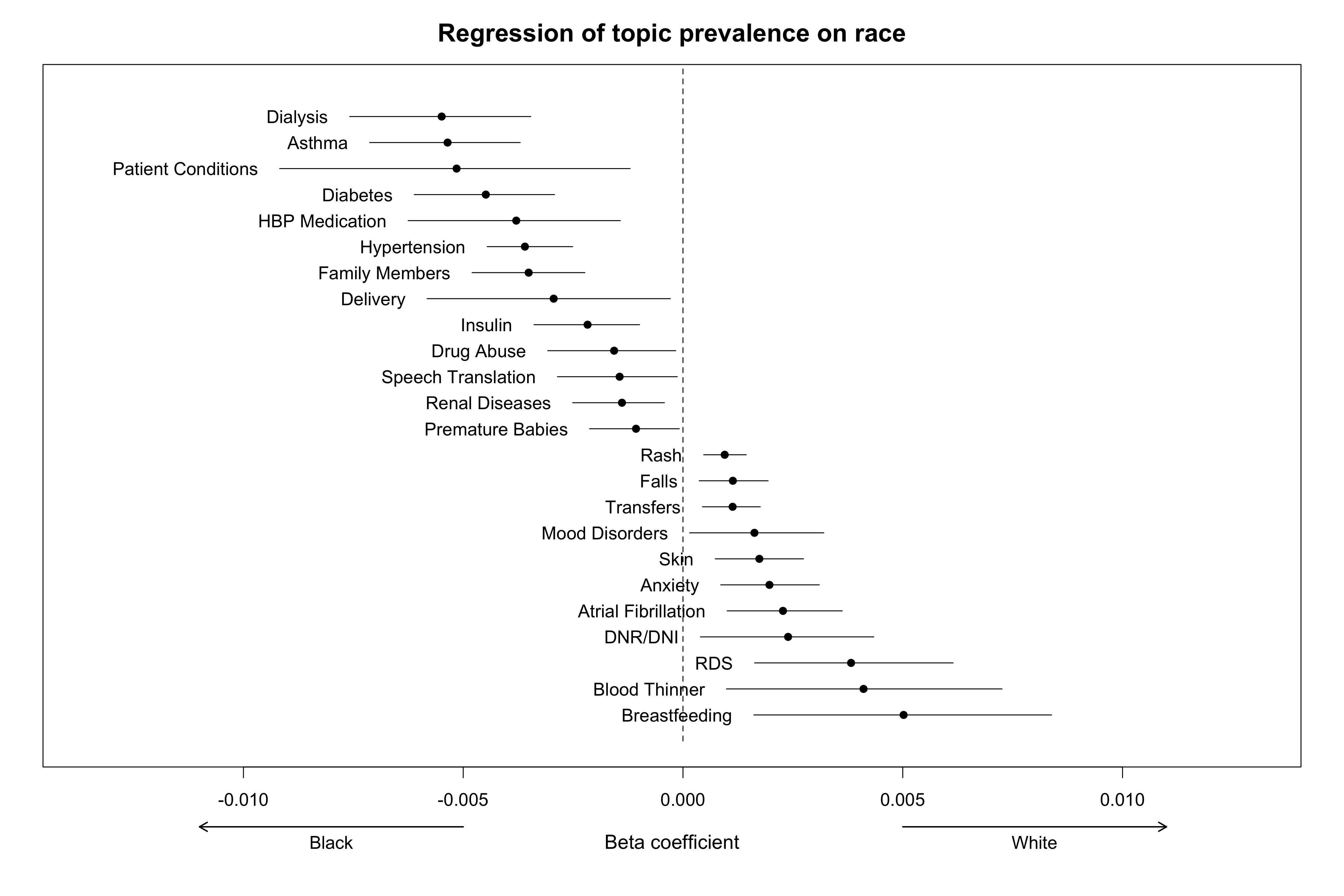}
    \caption{Differences in the topics discussed in notes written for Black and White patients in MIMIC. Topics were algorithmically identified by running a structural topic model (STM) on the MIMIC notes with k=200 topics and race as a covariate. Topics to the left of the dashed vertical line are significantly more common in notes written for Black patients, while those to the right are significantly more common in notes written for White patients. The chart plots the mean effect of race on topic prevalence with an error bar signifying the 95\% confidence interval (note that we only display the subset of topics with significant effects). Topics were manually labeled using the high-probability words identified by the STM.}
    \label{fig:stm}
\end{figure*}

We summarize these more nuanced differences in content between notes written for White and Black patients using a structural topic model (Figure \ref{fig:stm}). This analysis largely supports existing findings around comorbidities and skin. However, it yields a few additional insights: for example, discussion of mental health conditions like anxiety is much more common in the clinical notes of White patients. This may again reveal a systemic issue, as anxiety disorders are understudied, underdiagnosed, and undertreated in Black populations \cite{57-Vanderminden2019-xz, 61-Williams2013-ev}. Overall, the STM establishes that there are several implicit indicators of race in nursing notes, which makes redacting patient race a challenging task.

\subsubsection{Predictors of Race Are Deep-rooted in the Text}

While we have focused on the top predictors of patient self-reported race, we also find that the ensemble model is able to perform well over chance even after removing the strongest predictors of race from the MIMIC notes (Table \ref{tab:removing features auc}). This finding indicates that the signals of race are deeply rooted in the text, and simply removing some words will not address this issue. 

\begin{table}[!b]
\centering
\caption{Ablation results for the ensemble model in the MIMIC dataset. Removing the top 25 most predictive words for each race (according to logistic regression coefficients) impacts performance, but the model is still able to detect race.}
\label{tab:removing features auc}
\resizebox{.75\linewidth}{!}{%
\begin{tabular}{@{}ll@{}}
\toprule
                                      & AUC  \\ \midrule
With all features                     & 0.83 \\
Removing common skin-related features & 0.76 \\
Removing top 25 features               & 0.73 \\ \bottomrule
\end{tabular}%
}
\end{table}

\section{Race Detection by Human Experts}

In the previous section, we established that ML models can infer patient race from nursing notes that are stripped of explicit racial identifiers. We further identified a number of race-based differences in clinical notes that drive this predictive performance. In this section, we evaluate whether humans are also able to identify patient self-reported race from redacted clinical notes. In a survey of 42 physicians, we found this is not true. The surveyed physicians not only struggled to accurately predict patient race, but often admitted that their predictions were no better than complete guesses. This finding speaks to the limits of human supervision of ML models: if a model were relying on its covertly inferred estimate of patient race, human experts would likely not be able to tell.

\subsection{Methods}

We recruited 42 physicians via email to participate in a short web based experiment. This study was exempt from a full IRB ethical review, as it met the criteria for exemption defined in Federal regulation 45 CFR 46. We chose physicians as experts in this setting as they are both experienced in reading nursing notes, and a step removed from actually writing them. 

Consenting participants were shown ten notes from the MIMIC dataset. The selected notes were racially balanced (five White patients, five Black patients), and conveyed different levels of model accuracy: four notes that were predicted correctly, four that were predicted incorrectly, and two that the model was unsure about (i.e. $\sim50\%$ predicted probability of the patient being Black). For each note, participants were asked to indicate (1) whether they believed the patient was White/Caucasian or Black/African-American, and (2) how sure they were in their belief on a scale of 1-5, where 1 reflects a complete guess and 5 a strong belief. Participants were also given the option to highlight parts of the text that informed their belief.

We evaluated physicians on overall accuracy (i.e. the percentage of patients whose race they identified correctly), sensitivity for Black patients (i.e. the number of Black patients who were correctly identified as being Black), and the positive predictive value (PPV) for White patients (i.e. the number of predicted White patients who were actually White). We also evaluated the agreement between physician predictions using Fleiss’ kappa measure \cite{18-Fleiss1973-sm}.

\subsection{Results}

\begin{figure*}
    \centering
    \includegraphics[width=0.8\textwidth]{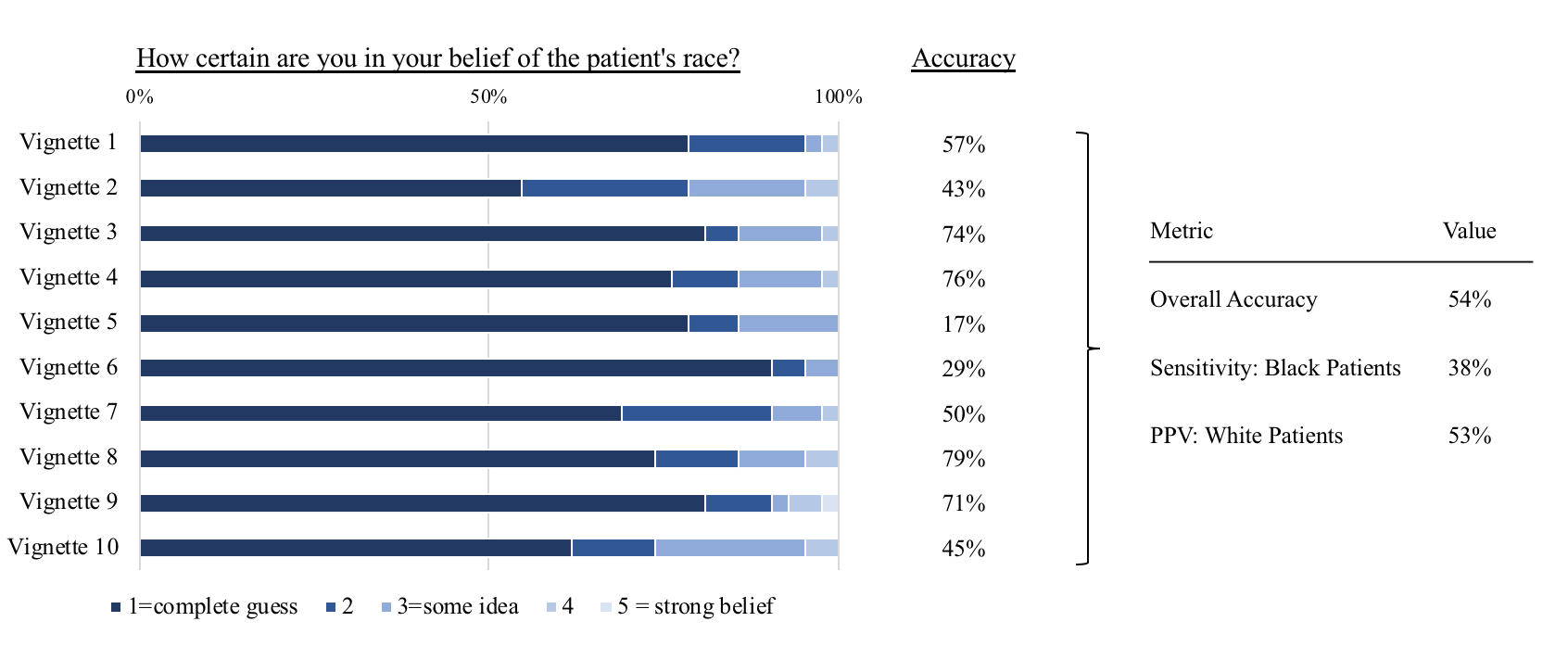}
    \caption{Assessing the ability of human experts to detect race from nursing notes. For all ten notes presented, the majority of the 42 surveyed physicians indicated that their prediction of the patient’s race was a complete guess. This lack of surety is borne out in their predictions, which are barely better than chance (average accuracy of 54\% across vignettes). The physicians have low sensitivity in identifying Black patients (38\%) and low positive predictive value (PPV) in identifying White patients (53\%), indicating that they may be defaulting to guessing a patient is White.}
    \label{fig:physcians}
\end{figure*}

We found that the physicians in our study were unable to predict patient race, with an average accuracy of $54\%$ ($n=420$ responses), only slightly better than chance (Figure \ref{fig:physcians}). While they were more accurate for White patients ($70\%$ vs $37\%$ for Black patients), this is likely as they defaulted to guessing a patient was White - the positive predictive value for White patients is just $53\%$. There was also only slight agreement between physicians, with their predictions exhibiting a Kappa statistic of $0.05$ (rejected null hypothesis of no agreement with z-value=$4.84$, p-value$< 0.001$).

Crucially, in a vast majority of cases ($\sim75\%$), physicians indicated that their prediction was a complete guess. Moreover, accuracy did not increase with self-reported certainty; physicians who said they had “some idea” of a patient’s race were less accurate than the average respondent (40\%, $n=43$ responses). 

\section{Simulation Experiment on Bias Propagation}

While it is not inherently problematic for a model to be able to predict a patient’s race from nursing notes, it is concerning if these differences lead to poorer performance. We perform an experiment using actual clinical notes with a synthetically generated biased treatment decision. We show that if ML models are trained on biased decisions, they make biased recommendations even without explicit access to patient race. 

\subsection{Methods}

We demonstrate the dangers of race-inferring models through a simulation experiment. Prior work has established the existence of several racial disparities in clinical treatment decisions. For example, Black patients are $\sim30\%$ less likely to be prescribed analgesia for acute pain in emergency settings than White patients \cite{33-Lee2019-bf}. Black patients are also less likely than White patients to be given appropriate cardiac care \cite{7-Arora2018-ym}, to receive kidney dialysis or transplants \cite{41-Ng2020-wk}, and to receive the best treatments for stroke, cancer \cite{28-Hershman2005-gl}, and AIDS \cite{40-Nelson2002-gf}. Our simulation evaluates whether ML models can perpetuate such biases in treatment even if they are trained on race-redacted clinical notes.

Our experiment uses real clinical notes from MIMIC with a synthetically generated, biased treatment decision. Because White patients far outnumber Black patients in MIMIC (Table \ref{tab:s1-demographic}), we created a balanced dataset of 2,014 adult Black patients and 2,014 adult White patients with random undersampling. Note that this approach captures all the adult Black patients in our cohort. We assumed that 50\% of these patients had a clinical condition (e.g. acute pain) that made them eligible for a specific treatment (e.g. analgesia). The presence of this condition was randomly assigned so that it was equally prevalent in Black and White patients. However, the decision to administer treatment to a patient with this condition was racially biased, that is, the treatment was assigned at a higher rate to White patients than Black patients. This biased decision resembles previously discussed disparities in analgesia prescription and other clinical care \cite{33-Lee2019-bf, 40-Nelson2002-gf}.

We then evaluated whether a model trained to make this decision would perpetuate the treatment gaps in the data. We trained an L2-penalized logistic regression \cite{46-Pedregosa2011-yz} to predict the treatment from a patient’s nursing notes, using an 8:2 train-test split. As before, we used a unigram BoW representation of the nursing notes. These notes were race-redacted, that is, contained no explicit identifiers of patient race. After removing stop words, tokenizing, and lemmatizing, the final vocabulary consists of 54,432 words. The model also received the presence of the clinical condition as an additional variable. We evaluated our model on the test set, and assessed whether the trained model was significantly less likely to recommend the treatment to Black patients than White patients. If such a gap exists, then the bias from the training data has propagated to the model, as differential treatment rates by race create differential model recommendations. We assessed this bias propagation for various magnitudes of training bias (10-50\%), and report average results and 95\% confidence intervals across 100 simulations.

\subsection{Results}

\begin{figure}
    \centering
    \includegraphics[width=\linewidth]{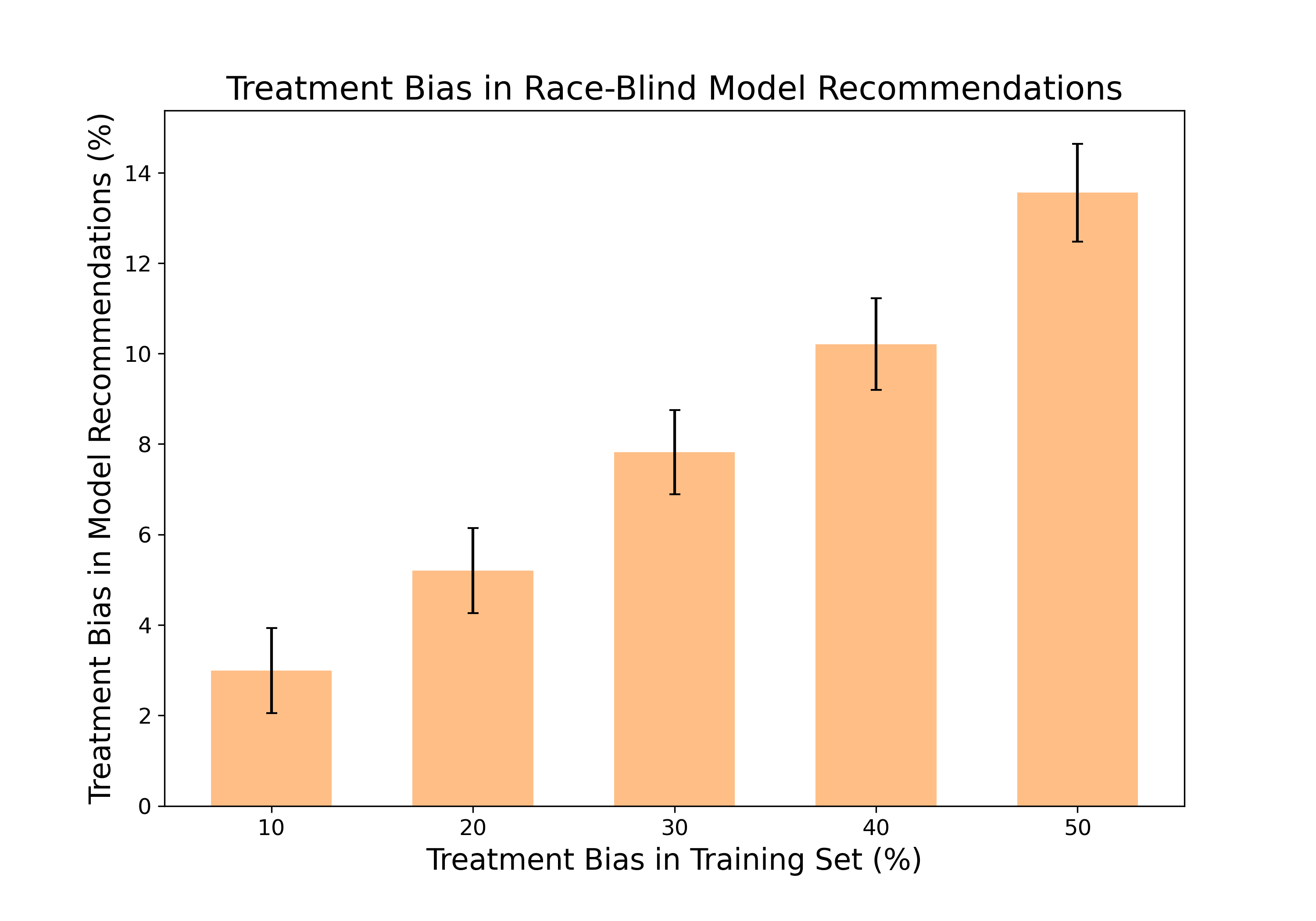}
    \caption{The mean recommendation bias at each level of treatment bias, along with an error bar that signifies the 95\% confidence interval. The x-axis plots the racial bias in the treatment decision. For example, a 10\% treatment bias describes a situation in which  80\% of eligible White patients were administered the treatment, but only 70\% of eligible Black patients were. The y-axis plots the corresponding bias in model predictions: how much less likely was the model to recommend the treatment for a Black patient than a White patient?}
    \label{fig:bias}
\end{figure}

We found that even without access to patient race, the model propagates the bias in the training data, and is significantly less likely to recommend the treatment to Black patients (Figure \ref{fig:bias}). This trend is observed for both small ($\leq20\%$) and large ($\geq30\%$) levels of training bias. We find that the level of propagation scales with the level of induced disparity, i.e, a training set disparity of 10\% results in a 3\% gap in model recommendations, while a 30\% disparity creates nearly a 10\% gap. The magnitude of the training set bias may be generally reduced in the model recommendation gaps because models do not predict race perfectly. However, the bias is replicated here with only the redacted notes as data, and no direct access to patient self-reported race or any other correlated demographics. While this finding is perhaps not surprising, it has not been noted before in prior work using clinical notes, and is important to highlight given the severe consequences of undetected bias propagation. 

Overall, our treatment simulation experiment demonstrates that even if a note contains no explicit information on patient race, the implicit racial information provides a signal that ML models could use to propagate existing biases in clinical care. We know from prior work that the absence of racial information in data is a sufficient condition for achieving fairness in machine learning recommendations \cite{38-Madras2018-pj}. However, other work has shown that learning race-blind representations is challenging, and biases are hard to remove through standard adversarial techniques \cite{64-Zhang2020-ze}. Our work further emphasizes this fact: racial information is deeply rooted in clinical notes, and implicit signals provide a potential vector for bias propagation. 

\section{Discussion and Conclusion}

Our work demonstrates that models are able to accurately predict patient self-reported race in the redacted notes of Black and White patients, where human experts are not. We also simulate the implications of this difference in practice, given a biased treatment setting. Our work has several key implications for clinical practice and the deployment of ML tools in healthcare settings.

First, our work highlights potential areas of missed care. The investigation of our model's performance revealed some differences in clinical notes that were hard to explain; for example, references to bruising or rashes are extremely predictive of White race, even though there is no clinical reason for these symptoms to be less common in Black patients. While our analysis is not sufficient to establish that nurses are missing skin symptoms in Black patients, the strong association between these terms and White race does suggest the possibility of missed care. Our findings are very concerning from a clinical perspective since patients in the ICU setting are often non-mobile and are at greater risk for skin damage and underlying soft tissue breakdown. These preventable injuries often cause pain, infection and patient harm \cite{44-Oozageer_Gunowa2018-jn}, and other clinicians have noted the need for increased education to identify skin damage in darker skin to avoid harmful consequences \cite{37-Lyder2009-wq, 43-Oozageer_Gunowa2020-dp}. Another concerning observation is the more frequent use of words like “demanding” and “difficult” for Black patients, which may hint at differential treatment. Investigating these trends further and causally establishing the presence of disparities in clinical care based on differences in documentation is an important avenue for future work.

Second, the risk of bias propagation is compounded by the fact that human experts do not share the ability to identify race from clinical notes. This finding establishes the limits of human oversight of ML systems \cite{32-Koulu2020-vp, 50-Robinette2017-mt}. Standard machine learning interpretability techniques highlight important features used by models in making predictions \cite{22-Gilpin2018-yn}. Even if these techniques worked perfectly, human experts would not be able to judge whether highlighted predictors were implicitly conveying racial information. Thus, if a model inferred race in making clinical predictions, human experts may not be able to detect this racial bias. This emphasizes the need to explicitly incorporate fairness considerations when designing ML systems in healthcare. It is vital to embed automated fairness checks and constraints \cite{1-Adebayo2016-ht, 3-Aggarwal2019-qk, 11-Bellamy2019-rw} at every stage in the ML pipeline, from data collection \cite{15-Chen2018-ka} to algorithm development \cite{38-Madras2018-pj} to deployment \cite{16-Chen2021-uv}. 

Finally, we emphasize that removing explicit racial identifiers from clinical notes is not sufficient to obscure patient race. This finding is vital to consider when designing algorithms to support clinical decision making. As our simulation experiment demonstrates, algorithms can still propagate existing biases in clinical care even if trained in a seemingly race-blind fashion. The combination of existing health disparities and potentially race-inferring algorithms makes it incredibly easy to unintentionally encode disparate treatment. Any ML model trained on clinical notes must thus be thoroughly and continuously audited for racial bias both before and after deployment \cite{21-Ghassemi2022-dd, 47-Raji2020-lb, 59-Wiens2019-we}.

\begin{acks}
This work is supported by the MIT-IBM Watson AI Lab. HA is funded by the MIT Jameel Clinic. KC is funded by the National Institute of Nursing Leadership through grant R01NR016941-01 Communicating Narrative Concerns Entered by RNs (CONCERN). LAC is funded by the National Institute of Health through the NIBIB R01 grant EB017205. MG is funded by the CIFAR Azreili Global Scholar and the Helmholtz Professorship.

\end{acks}

\bibliographystyle{ACM-Reference-Format}
\bibliography{dtn_references}

\clearpage

\setcounter{table}{0}
\renewcommand{\thetable}{S\arabic{table}}

\appendix
\section{Descriptive statistics for patients}
\vspace{-1in}
\begin{table}[!htp]
\centering
\caption{Descriptive statistics of the patient cohort in each dataset, including demographics, insurance provider, unit type, and comorbidities. We report the number of patients in each category with the percentage of total patients in parentheses. }
\label{tab:s1-demographic}
\resizebox{\linewidth}{!}{%
\begin{tabular}{@{}llrr@{}}
\toprule
                        &              & \multicolumn{2}{c}{Dataset}   \\
                        &              & MIMIC         & Columbia      \\ \midrule
Total                   &              & 28,032        & 29,807        \\
                        &              &               &               \\
Race                    & Black        & 2,833 (10\%)  & 5,883 (20\%)  \\
                        & White        & 25,199 (90\%) & 23,924 (80\%) \\
                        &              &               &               \\
Gender                  & Male         & 15,594 (56\%) & 16,696 (56\%) \\
                        & Female       & 12,438 (44\%) & 13,110 (44\%) \\
                        &              &               &               \\
Insurance               & Public       & 15,195 (54\%) & 19,206 (64\%) \\
                        & Private      & 12,562 (45\%) & 10,158 (34\%) \\
                        & Self-Pay     & 275 (1\%)     & 443 (1\%)     \\
                        &              &               &               \\
Age                     & \textless{}1 & 5,495 (20\%)  & 3,728 (13\%)  \\
                        & 1-17         & 0 (0\%)       & 35 (0\%)      \\
                        & 18-24        & 623 (2\%)     & 545 (2\%)     \\
                        & 25-34        & 902 (3\%)     & 1,286 (4\%)   \\
                        & 35-44        & 1,725 (6\%)   & 1,553 (5\%)   \\
                        & 45-54        & 3,176 (11\%)  & 2,998 (10\%)  \\
                        & 55-64        & 4,332 (15\%)  & 5,280 (18\%)  \\
                        & 65-74        & 4,497 (16\%)  & 6,489 (22\%)  \\
                        & 75+          & 7,282 (26\%)  & 7,893 (26\%)  \\
                        &              &               &               \\
Unit                    & MICU         & 8,763 (31\%)  & 5,770 (19\%)  \\
                        & NICU         & 5,495 (20\%)  & 3,767 (13\%)  \\
                        & SICU         & 4,247 (15\%)  & 3,583 (12\%)  \\
                        & CCU          & 3,805 (14\%)  & 5,175 (17\%)  \\
                        & TSICU        & 3,334 (12\%)  & 0 (0\%)       \\
                        & CSRU         & 4,761 (17\%)  & 7,145 (24\%)  \\
                        & NUICU        & 0 (0\%)       & 4,367 (15\%)  \\
                        &              &               &               \\
Number of Comorbidities & 0            & 7,023 (25\%)  & 4,135 (14\%)  \\
                        & 1            & 3,614 (13\%)  & 2,533 (8\%)   \\
                        & 2            & 4,682 (17\%)  & 3,459 (12\%)  \\
                        & 3            & 4,513 (16\%)  & 4,031 (14\%)  \\
                        & 4            & 3,470 (12\%)  & 4,236 (14\%)  \\
                        & 5            & 2,303 (8\%)   & 3,652 (12\%)  \\
                        & 6            & 1,266 (5\%)   & 2,738 (9\%)   \\
                        & 7+           & 1,161 (4\%)   & 5,023 (17\%)  \\ \bottomrule
\end{tabular}%
}
\end{table}

\section{Words Removed From the Columbia Notes}
\begin{table}[H]
\centering
\caption{A list of PHI removed from the Columbia notes to redact race. These terms often served as strong proxies for race, and were thus removed.}
\label{tab:words removed columbia}
\resizebox{.4\linewidth}{!}{%
\begin{tabular}{@{}cl@{}}
\toprule
Area Codes & Places / Hospitals    \\ \midrule
347       & brooklyn   \\
646       & harlem     \\
201       & interfaith \\
845       & olmstead   \\
908       & downstate  \\
430       & 5gn        \\
185       & cosgrove   \\
718       & zaire      \\
          & africa     \\
          & jamaica    \\
          & regional   \\
          & samaritan  \\
          & valley     \\ \bottomrule
\end{tabular}%
}
\end{table}

\section{Bigram Representation}
\label{sec:bigram}

In addition to a unigram BoW representation of the clinical notes, we also tested models that used a unigram + bigram BoW representation. The results of these models on the MIMIC dataset are provided in Table \ref{tab:bigram}. As these models did not meaningfully improve performance, we restricted our focus in the main paper to unigram only models. 

\begin{table}[H]
\centering
\caption{Classification accuracy of bigram representations. We found that adding bigrams to the BoW representation did not improve accuracy. As before, we evaluated the classifier on ten random train-test splits, and report the mean and standard deviation of test set AUCs across
splits. }
\label{tab:bigram}
\resizebox{.8\linewidth}{!}{%
\begin{tabular}{@{}llc@{}}
\toprule
BoW Representation & Model & AUC    \\ \midrule
Unigram & Logistic Regression & 0.78 (0.004) \\
 & xgBoost & 0.81 (0.003) \\
 & Ensemble & 0.83 (0.003) \\
Unigram + Bigram & Logistic Regression &  0.78 (0.005) \\
 & xgBoost & 0.82 (0.004) \\
 & Ensemble & 0.83 (0.003) \\
          \bottomrule
\end{tabular}%
}
\end{table}



\end{document}